\documentclass{article}

\usepackage{arxiv}

\usepackage[utf8]{inputenc} 
\usepackage[T1]{fontenc}    
\usepackage{hyperref}       
\usepackage{url}            
\usepackage{booktabs}       
\usepackage{amsfonts}       
\usepackage{nicefrac}       
\usepackage{microtype}      
\usepackage{lipsum}		
\usepackage{graphicx}
\usepackage{natbib}
\usepackage{doi}
\usepackage{threeparttable} 
\usepackage{amsmath,bm} 
\usepackage{bbm} 
\usepackage{subcaption} 
\usepackage{derivative} 

\title{The impact of heteroskedasticity on uplift modeling}

\author{ \href{https://orcid.org/0000-0000-0000-0000}{\includegraphics[scale=0.06]{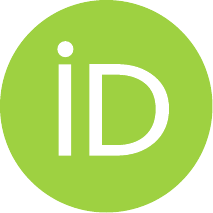}\hspace{1mm}Björn Bokelmann} \\
	Chair of information systems\\
	Humboldt University Berlin\\
	Unter den Linden 6\\
    10099 Berlin\\
	\texttt{bokelmab@hu-berlin.de} \\
	\And
	\href{https://orcid.org/0000-0000-0000-0000}{\includegraphics[scale=0.06]{orcid.pdf}\hspace{1mm}Stefan Lessmann} \\
	Chair of information systems\\
	Humboldt University Berlin\\
	Unter den Linden 6\\
    10099 Berlin\\
}

\renewcommand{\shorttitle}{Heteroscedasticity in uplift modeling}

\hypersetup{
pdftitle={Binarized target modeling},
pdfsubject={q-bio.NC, q-bio.QM},
pdfauthor={Björn~Bokelmann, Stefan~Lessmann},
pdfkeywords={First keyword, Second keyword, More},
}

\begin{document}
\maketitle

\begin{abstract}
There are various applications, where companies need to decide to which individuals they should best allocate treatment. To support such decisions, uplift models are applied to predict treatment effects on an individual level. Based on the predicted treatment effects, individuals can be ranked and treatment allocation can be prioritized according to this ranking. An implicit assumption, which has not been doubted in the previous uplift modeling literature, is that this treatment prioritization approach tends to bring individuals with high treatment effects to the top and individuals with low treatment effects to the bottom of the ranking. In our research, we show that heteroskedastictity in the training data can cause a bias of the uplift model ranking: individuals with the highest treatment effects can get accumulated in large numbers at the bottom of the ranking. We explain theoretically how heteroskedasticity can bias the ranking of uplift models and show this process in a simulation and on real-world data. We argue that this problem of ranking bias due to heteroskedasticity might occur in many real-world applications and requires modification of the treatment prioritization to achieve an efficient treatment allocation.
\end{abstract}


\section{Introduction}
There are various applications, where companies need to decide to which individuals they should best allocate treatment: Online marketing companies decide which costumers to provide coupons, in order to increase their spendings \citep{haupt2022targeting}, telecomunication companies aim to provide offers to customers to prevent them from churning \citep{wagh2023customer} and advertisers try to select the right advertisement for individuals in the internet to generate clicks \citep{ke2021addressing}. These companies frequently use data-driven ways to identify on which individuals a treatment has the highest positive effect. The treatment effect of individuals is called \textit{conditional average treatment effect (CATE)}. Predicting the CATE of an individual is called \textit{uplift modeling}. Those uplift models are build on historical data containing individuals characteristics (features), treatment status and outcome of interest. They are applied to rank individuals by the predicted treatment effect, such that the high-ranked individuals can be prioritized for the treatment. 

In previous literature there has been relatively little concern, whether this approach of uplift modeling and treatment prioritization really works as expected. Can we really trust uplift models to bring individuals with high treatment effects to the top of the ranking and individuals with low treatment effects to the bottom? In our research, we show that statistical properties of the training data can cause a bias of the uplift model ranking: individuals with the highest treatment effects can get accumulated in large numbers at the bottom of the ranking. With this bias in the ranking, the common principle of allocating treatment to the highest ranked individuals and not treating individuals at the bottom of the ranking becomes inefficient. 

The statistical property, which can cause this uplift model ranking bias is called \textit{heteroskedasticity}. Data is said to be heteroskedastic, if the magnitude of noise in the outcome of individuals is heterogenous and depends on the features of the individuals. In the statistics and econometrics literature, heteroskedasticity is a topic of high importance, reflected in a large number of studies about it.\citep{white1980heteroskedasticity,engle1982autoregressive,park1966estimation} In contrast, in uplift modeling, the problem of heteroskedasticity has not yet caught attention and the potential problem of a ranking bias is still unknown.   

In this paper, we show that heteroskedasticity tends to create heterogenous uplift model variance, whereby the uplift model variance increases with the magnitude of outcome noise. We then show, that, due to the heterogeneity of model variance, individuals with high outcome noise tend to get ranked at the bottom or top by uplift models, irrespective of their actual CATE values. Finally, we derive a condition for binary outcomes, under which this heteroskedasticity effect can lead to a ranking bias, with individuals having high treatment effects beeing accumulated at the bottom in large numbers. The condition causing this bias is that individuals with the highest probabilities of positive outcomes tend to have the highest CATE. We argue why this might be the case in many real-world applications and show this condition with the resulting ranking bias on a real-world data set.

In summary, we provide solid evidence of a potential problem in uplift modeling, which can limit the the application benefit of this approach. We close the paper by discussing potential solutions to the problem, which could be the topic of further research.

\section{Related literature}
The estimation of CATE, is the topic of two different literature branches. The first literature branch is called \textit{heterogenous treatment effect} (HTE) estimation. Thereby HTE has the same meaning as CATE. The HTE estimation literature is more theory and less application oriented. It is part of the literature about statistics and econometrics. The focus of this literature is to provide CATE estimation methods fulfilling favorable statistical properties (consistency, efficiency, valid inference) The second literature branch is called \textit{uplift modeling}. The uplift modeling literature is less theory-driven and more application oriented. It evolved from the problem of efficient treatment allocation in applications like online marketing or costumer churn prevention, where the task is to identify a subset of individuals, on which the treatment yields the highest beneficial effect. Instead of pursuing statistical properties, the focus of this literature is to provide the optimal treatment prioritization, according to economic measures. Both literature branches apply similar methods, including the two-model approach (T-learner) \citep{hansotia2002incremental}, the single-model approach (S-learner) \citep{kunzel2019metalearners} and the causal random forest \citep{athey2019generalized}. In the following, we review to what extend these literature branches have dealt with the problem of heteroskedasticity for treatment prioritization.

In the statistics and econometrics literature, heteroskedasticity has been acknowledged as a problem which can render ordinary lest squares regression-based average effect estimates inefficient and render inference methods, like confidence intervals and statistical tests invalid. For models based on maximum likelihood estimation, heteroskedasticity can even render effect estimates biased.\citep{greene2000econometric} To deal with the problem of heteroskedasticity, various statistical methods have been introduced, including heteroskedasticity-robust standard error estimation \citep{white1980heteroskedasticity}, bootstrap-based inference \citep{davison1997bootstrap} and weighted least squares \citep{strutz2011data}. In contrast to regression-based estimation of average treatment effects, heteroskedasticity is less of a concern in the literature about HTE estimation. This is probably due to the fact, that most of the HTE estimation methods are based on flexible non-parametric models and do not require homoskedasticity to yield valid inference.\citep{wager2018estimation,athey2019generalized,chernozhukov2018double} In particular, there is no literature about the impact of heteroskedasticity in HTE estimation, when the application goal is treatment prioritization. This is not surprising, as treatment prioritization is less of a focus in the HTE estimation literature, reflected by the fact that metrics evaluating treatment prioritization are rarely used.

The uplift modeling literature, in contrast, predominantly applies ranking metrics \citep{radcliffe2007using} or metrics calculating a monetary value of the treatment prioritization \citep{gubela2020response,haupt2022targeting,verbeke2023or}, to evaluate models. Related to this focus on ranking and treatment prioritization, the uplift modeling literature provides methods specifically optimized for ranking (in contrast to precise CATE estimation), under the banner "learning to rank for uplift modeling".\citep{betlei2021uplift,devriendt2020learning} However, heteroskedasticity does not play a role in these studies. The study which provides results closest to our findings is conducted by \citet{huang2022less}. The authors identify a "misattribution problem" when applying uplift models for treatment prioritization. They find that the treatment effect of low-ranked individuals gets systematically underestimated and the effect of high-ranked individuals gets systematically overestimated. They also find, that the reason for this misattribution problem is the unexplained outcome variance (noise). This is in-line with our own findings. However, what \citet{huang2022less} do not consider is the link to heteroskedasticity. By leaving out the aspect of heteroskedasticity, unexplained variance is treated is an overall problem, negatively impacting the treatment prioritization and the solution proposed is a method to reduce this unexplained variation (which works, if the outcome is a long term event). By considering heteroskedasticity, we can explain in detail which individuals are likely to be misattributed to the bottom or top of the ranking and, whats more, we can show that the problem due to the unexplained variance can be even much more grave than effect under- or overestimation at the top or bottom of the ranking. Heteroskedasticity can even systematically bias the ranking so far, that treatment allocation according to the ranking becomes inefficient and it would be better to treat the highest and lowest ranked individuals instead.

\section{The impact of heteroskedasticity on uplift modeling}
\subsection{Heteroskedasticity}
We consider the case, where we have an outcome $y$, feature values $x$ and treatment status $w$, which is assigned randomly with probability $p$ for treatment. The response can then be written as
\begin{align*}
    y=\mu_x+w\cdot\tau_{x}+\varepsilon,
\end{align*} where $\mu_x:=E[y|x,w=0]$ is the conditional expected value of the outcome and $\tau_x:=E[y|x,w=1]-E[y|x,w=0]$ is the CATE. In this paper, the unexplained variance of the outcome (also called noise) is of high interest. For a binary outcome, the unexplained variance for untreated individuals with features $x$ is given by 
\begin{align*}
    Var[y|x,w=0]&=\mu_x\cdot (1-\mu_x).
\end{align*} So, the magnitude of unexplained variance in the outcome of an individual depends on the conditional outcome probability $\mu_x$. An individual with $\mu_x$ close to zero or one has very little unexplained variance in the outcome. For those individuals it is quiet certain that the outcome will be zero respectively one. In contrast, for individuals with $\mu_x$ close to 0.5, there is high unexplained variance in the outcome. For those individuals it is very uncertain whether the outcome will be zero or one.

If the conditional outcome probability $\mu_x$ is heterogeneous between the individuals, so is the magnitude of noise. Such heterogeneity in the noise with regard to the features $x$ is called \textit{heteroskedasticity}. We are interested in such heteroskedasticity, because we hypothesize the following impact: $Var[y|x,w=0]$ is a factor, which, in addition to $\tau_x$, impacts the ranking of an individual with features $x$. The Uplift models tend to rank individuals with a high $Var[y|x,w=0]$ at the bottom or top, while individuals with a low $Var[y|x,w=0]$ tend to be ranked in the middle.

\subsection{Heterogenous model variance}
To describe the impact of heteroskedasticity on uplift modeling, we need to analyze uplift modeling with regard to bias and variance. The predictions of an uplift model could be expressed by
\begin{align}
    \hat{\tau}=\tau_x+\varepsilon_{\hat{\tau}}. \label{eq_rep_pred}
\end{align} In case the uplift model is unbiased, the error term fulfills $E[\varepsilon_{\hat{\tau}}|x]=0$. Thereby, $\varepsilon_{\hat{\tau}}$ is a random variable, which represents the randomness in the model building procedure. The model variance for given features $x$ is then 
\begin{align*}
    Var[\hat{\tau}|x]=Var[\varepsilon_{\hat{\tau}}|x].
\end{align*} We hypothesize that this model variance depends on the unexplained outcome variance: The higher the unexplained outcome variance for an individual with features $x$, the higher the model variance tends to be. 

To examine our hypothesis. We simulate a randomized controlled trial, with treatment probability of 50\%. We only simulated one feature $x$, which creates heterogeneity in the outcome probability $\mu_x$, but the CATE is constantly zero for all individuals. In total, we simulated 10,000 data sets with 2,002 observations of the form:
\begin{align*}
    x&\sim U_{[0,1]} \text{ (one feature with uniform distribution)}\\
    \mu_x&=x \\
    \tau_x&=0 \\
    w&\sim Bern(0.5)\\
    y&\sim Bern(\mu_x+w\cdot\tau_x)
\end{align*}
We applied three common uplift modeling approaches: the causal random forest (CRF) \citep{athey2019generalized}, the S-learner \citep{kunzel2019metalearners} and the T-learner \citep{hansotia2002incremental}.

\begin{figure}
\centering
\centering
  \includegraphics[width=.9\linewidth]{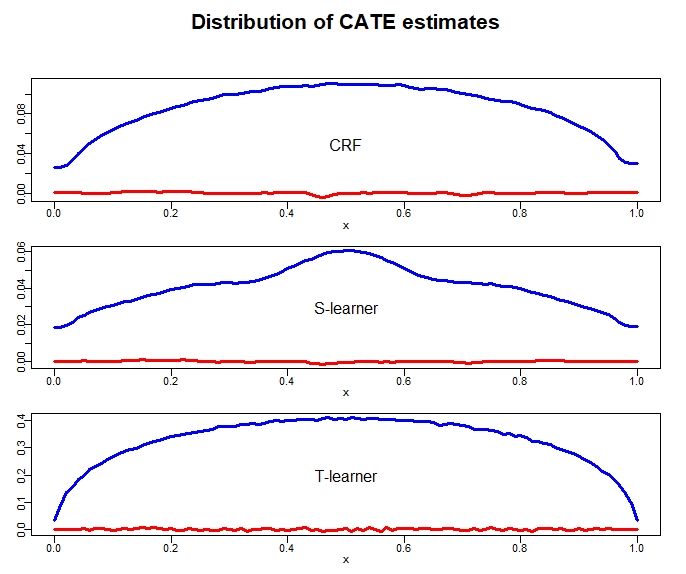}
  \captionof{figure}{Conditional distribution of uplift model predictions. The red line represent the expected value $E[\hat{\mu}|x]$ of the respective uplift models and the blue line represent the model standard deviation $\sqrt{Var[\hat{\mu}|x]}$ at point $x$.}
  \label{fig:het_var}
\end{figure}

The results of this simulation are illustrated in figure \ref{fig:het_var}. We can see, that the expected value of all the uplift modeling approaches is close to zero for each $x$. As we defined $\tau_x=0$ in the simulation, we can conclude that there is no sign of a bias. For the model standard deviation $\sqrt{Var[\hat{\mu}|x]}$, we can see an increase in $x$, for each $x<0.5$ and a decrease in $x$ for each $x>0.5$. This fits exactly to our hypothesis: As we defined $\mu_x=x$, the magnitude of unexplained outcome variance $Var[y|x,w=0]=\mu_x\cdot (1-\mu_x)$ increases in $x$ for $x<0.5$ and decreases in $x$ for $x>0.5$. 

Certainly, this very simple simulation is no proof for the hypothesis, that the model variance increases with the magnitude of unexplained outcome variance. In real-world applications, we usually have multiple features and the feature-target relationship is far more complex than in our little simulation. However, we will also provide evidence in real-world data, that model variance increases with the magnitude of unexplained outcome variance. This little simulation can be seen as an illustration of the process.

\subsection{The impact of heteroskedasticity on the ranking}
In practice, uplift models are usually applied to rank individuals, such that treatment can be allocated to the highest ranked individuals. In this way, the applicant hopes to achieve the highest beneficial effect, while saving resources by sparing treatment for low ranked individuals. 

Using equation \eqref{eq_rep_pred}, we analyze which factors yield a high or low ranking of an individual with features $x$. The individuals with a high respectively low ranking are those, for which $\tau_x+\varepsilon_{\hat{\tau}}$ is high respectively low. Clearly, one factor which influences the ranking is then $\tau_x$. If $\tau_x$ is high, it is likely that $\tau_x+\varepsilon_{\hat{\tau}}$ takes a high value. This is not surprising, as otherwise the idea of uplift model-based treatment prioritization would be pointless. However, $\tau_x$ is not the only factor determining whether an individual tends to get a high or low ranking. A high value of $\varepsilon_{\hat{\tau}}$ can also lead an individual to the top of the ranking and a low value of $\varepsilon_{\hat{\tau}}$ could lead the individual to the bottom of the ranking. And this is where heterogeneity in the model variance affects the ranking of individuals: A high model variance $Var[\hat{\tau}|x]=Var[\varepsilon_{\hat{\tau}}|x]$ for an individual with features $x$ means that this individual has a high probability that $\varepsilon_{\hat{\tau}}$ either takes a very high or very low value. Hence, this individual will have a high probability to get pushed to the top or bottom of the ranking. In contrast, an individual with low model variance is unlikely to be pushed to the top or bottom of the ranking and will thus tend to be ranked somewhere in the middle.  

\begin{figure}
\centering
\centering
  \includegraphics[width=.9\linewidth]{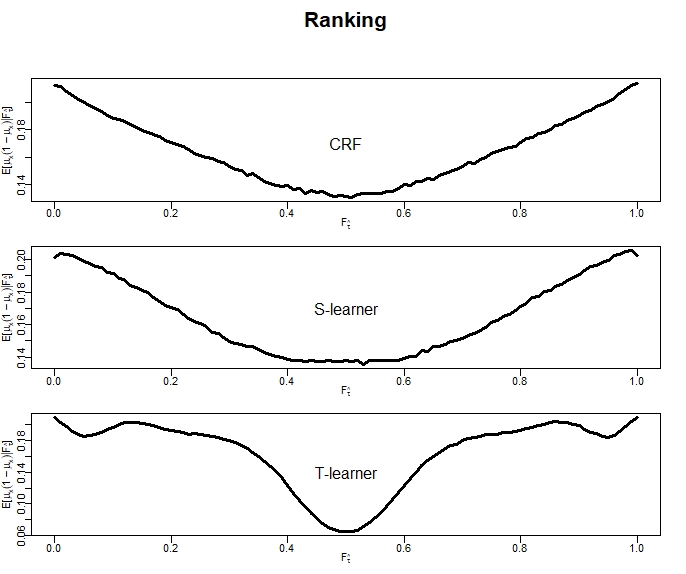}
  \captionof{figure}{Expected magnitude of noise $E[\mu_x(1-\mu_x)|F_{\hat{\tau}}]$ at each quantile $F_{\hat{\tau}}$ in the ranking.}
  \label{fig:sim_ranking}
\end{figure}

To examine whether we can observe this effect of heterogeneous model variance on the ranking in practice, we take the prediction results from the simulation of the last subsection and analyze the ranking of individuals. In the last section, we have seen that the model variance grows with the magnitude of unexplained variance $\mu_x(1-\mu_x)$. Now, we examine whether observations with high model variance (and therefore a high value of $\mu_x(1-\mu_x)$) tend to be ranked at the top and at the bottom. To do this, we group the individuals in 100 buckets, according to their model predictions $\hat{\tau}$. The first group $\{x:F_{\hat{\tau}}(\hat{\tau}(x))\in [0,0.01]\}$ contains the 1\% individuals with the lowest predictions. The next group $\{x:F_{\hat{\tau}}(\hat{\tau}(x)))\in [0.01,0.02]\}$ contains the individuals with predictions between the 1\%- and the 2\%-quantile, and so on. Here $F_{\hat{\tau}}(.)$ is the empirical cumulative distribution function within each simulation run. In this way, each bucket represents a certain position in the ranking. Within each bucket, we calculate the average value $E[\mu_x(1-\mu_x)|F_{\hat{\tau}}(\hat{\tau}(x))\in [\frac{i}{100},\frac{i+1}{100}]]$ of the magnitude of unexplained variance. We plot the corresponding results in figure \ref{fig:sim_ranking}. The horizontal axis $F_{\hat{\tau}}$ represents a position in the ranking, with zero being the bottom and one being the top. The fact that $E[\mu_x(1-\mu_x)|F_{\hat{\tau}}]$ is lowest in the middle and increases to the bottom and top shows that individuals with a high magnitude of unexplained outcome variance tend to get ranked at the top or bottom. So, we observe exactly what we expected. 

\subsection{The possibility of ranking bias}
In the last sections, we argued that the outcome probability $\mu_x$ affects the ranking of individuals and provided some evidence by a simulation. The question is, whether this has relevant consequences for uplift modeling and treatment prioritization. The application goal of uplift models is to rank individuals according $\tau_x$, such that it is possible to achieve a maximum cumulative treatment effect with limited resources, by only treating the highest ranked individuals. This only works, if the individuals with a high treatment effect can reliably be found on top of the ranking and individuals with low treatment effect can reliably be found at the bottom of the ranking. A potential problem resulting from the heteroskedasticity effect is the following: If individuals with high treatment effects, also tend to have a high magnitude of unexplained variance, the heteroskedasticity effect increases the probability that they appear at the top or bottom of the ranking. An increasing probability for individuals with high treatment effects to appear at the bottom of the ranking is very problematic because this means a significant share of individuals with high treatment effects will appear at the bottom of the ranking and would consequently not receive any treatment if the treatment allocation is done by the uplift model.

To discuss this problem more formally, we consider for a ranking position $p\in [0,1]$ the expected treatment effect of individuals ranked to this position. Thereby, $p=0$ indicates the bottom and $p=1$ indicates the top of the ranking. Mathematically, this expected treatment effect can be expressed by
\begin{align*}
    E[\tau_x|F_{\hat{\tau}}(\hat{\tau})=p].
\end{align*} When applying an uplift model for treatment effect prioritization, one would implicitly assume that the expected effect is the higher the higher the ranking, which means $E[\tau_x|F_{\hat{\tau}}(\hat{\tau})=p]$ monotonically increases with $p$. If this is the case, we call a ranking procedure \textit{unbiased}. If a ranking procedure is unbiased, a treatment allocation according to this ranking procedure makes sense: If treatment is allocated to the highest ranked individuals, we would expect the highest treatment effect. In contrast, if $E[\tau_x|F_{\hat{\tau}}(\hat{\tau})=p]$ starts at some point to decrease with $p$ (so, the expected effect decreases with the ranking), we speak of a \textit{ranking bias}. If a ranking bias occurs, treatment allocation according to this ranking becomes problematic: we can no longer expect to achieve the highest treatment effects by treating the highest ranked individuals because it is possible that higher effects would be achieved at lower positions of the ranking. 

The potential problem of heteroskedasticity in uplift modeling is that it can lead to such a ranking bias. If the magnitude of unexplained variance tends to increase with the CATE, extremely low rankings become increasingly likely for individuals with high CATE.Accordingly, at some $p$ close to 0 (position at the bottom of the ranking), the effect $E[\tau_x|F_{\hat{\tau}}(\hat{\tau})=p]$ increases as $p$ approaches 0. We show such a phenomenon on some real-world data in the next section. 

One way to diagnose ranking bias is to examine the Qini curve. The Qini curve represents the following: If we allocate treatment to all individuals with a ranking higher or equal to $p$, we would treat a share of $(1-p)$ individuals and expect to achieve the cumulative treatment effect $E[\tau_x|F_{\hat{\tau}}(\hat{\tau})>p]\cdot (1-p)$. The qini curve represents this cumulative effect for each possible share $(1-p)$ of treated individuals. Hence, the expected value of the Qini curve is 
\begin{align*}
    f(1-p)=E[\tau_x|F_{\hat{\tau}}(\hat{\tau})\geq p]\cdot (1-p).
\end{align*} As we show in Appendix \ref{ap_auuc},  it holds
\begin{align*}
    f'(1-p)=E[\tau_x|F_{\hat{\tau}}(\hat{\tau})=p].
\end{align*} If there is no ranking bias, $f'(1-p)$ would be monotonically decreasing with the share $(1-p)$. This corresponds to a strictly concave expected Qini curve. In contrast, ranking bias would correspond to a break of monotony of $f'(1-p)$ and thus a break in the concavity of the Qini curve.   

\section{Evidence from a real-world data set}
\subsection{Uplift modeling problem}
As a real-world data set, we analyze the Criteo online marketing data.\citep{diemert2021large} It is randomized controlled trial data with 13,979,592 observations and a treatment probability of 85\%. The target variable of this data set is conversion. The data set has already been used in several uplift modeling studies. Of these studies, the paper by \citet{fernandez2022causal} provided an astonishing result: The authors applied an uplift model and an outcome model (which predicts the outcome $y$ and not the causal effect $\tau_x$). They ranked the observations according to the predictions of both models and compared which of the models provided the better treatment prioritization. Astonishingly, the treatment prioritization of the uplift model proved to be worse than the prioritization by the outcome model. This raises doubts about the usefulness of uplift models for treatment prioritization. The authors explained this result by the high model variance of the uplift model. We decided to perform an analysis similar to \citet{fernandez2022causal} but, in contrast to the authors, analyze the astonishingly poor performance of the uplift model in the context of heteroskedasticity.

Just like in the paper by \citet{fernandez2022causal}, we applied a T-learner $\hat{\tau}(x)$ and an outcome model $\hat{\mu}(x)$, which we trained on 10\% of the observations, tuned on a separate 10\% of the observations and evaluated on the remaining 80\% of the observations. The reason why we decided to use such a large share of observations for the evaluation is the high variance in uplift evaluation metrics, which requires a high number of test observations to provide reliable results.\citep{bokelmann2023improving} As we analyze the shape of the Qini curve as a means to diagnose ranking bias, it is most important to have enough test observations to rule out that any peculiarities in the shape of the curve are by chance. 

\subsection{Ranking bias}
\begin{figure}
\centering
\centering
 \includegraphics[width=.7\linewidth]{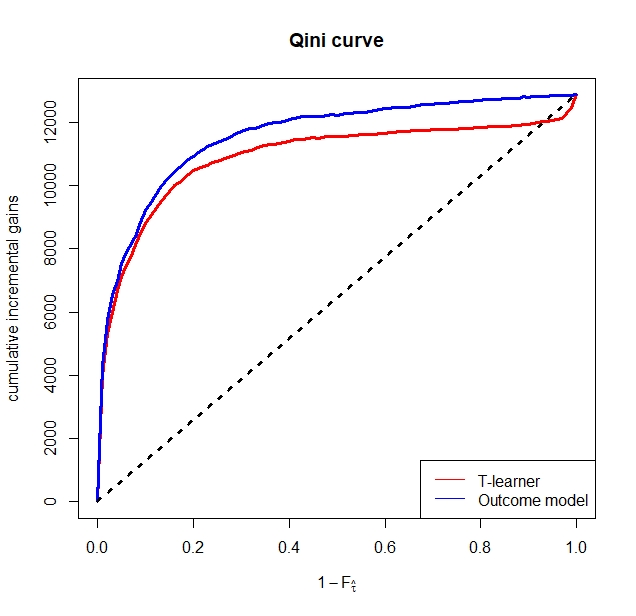}
 \captionof{figure}{Qini curves of the T-learner (red line) and the outcome model (blue line).}
 \label{fig:criteo_qini}
\end{figure}

The results of our analysis are provided in figure \ref{fig:criteo_qini}. Both the T-learner and the outcome model have Qini curves above the diagonal line, which shows that their ranking is better than a ranking by chance. Indeed, the outcome model has a higher Qini curve than the T-learner, which shows that its treatment prioritization is better. This is similar to the findings of \citet{fernandez2022causal}. However, what was not done by the authors was a detailed analysis of the poor performance of the T-learner. It is relatively easy to see, where the gap between both Qini curves stems from: The Qini curve shows a ranking bias with a break of concavity around the point 0.95 on the x-axis, while the outcome model has a completely concave Qini curve. The steep rise at the end of the Qini curve of the T-learner shows that these individuals should rather be ranked somewhere at the top. Due to the large sample size of the test set, it is certain that this pecularity in the shape is not by chance, but a solid evidence for ranking bias. 

With this ranking bias, it is clear that treatment prioritization by the ranking of the T-learner is inefficient. If only the highest ranked individuals would be chosen for treatment, a lot of potential would be lost. It would be better to allocate treatment to the highest ranked and the 5\% lowest ranked individuals.    

\subsection{Evidence of the heteroskedasticity impact}
\begin{figure}
\centering
\centering
 \includegraphics[width=.98\linewidth]{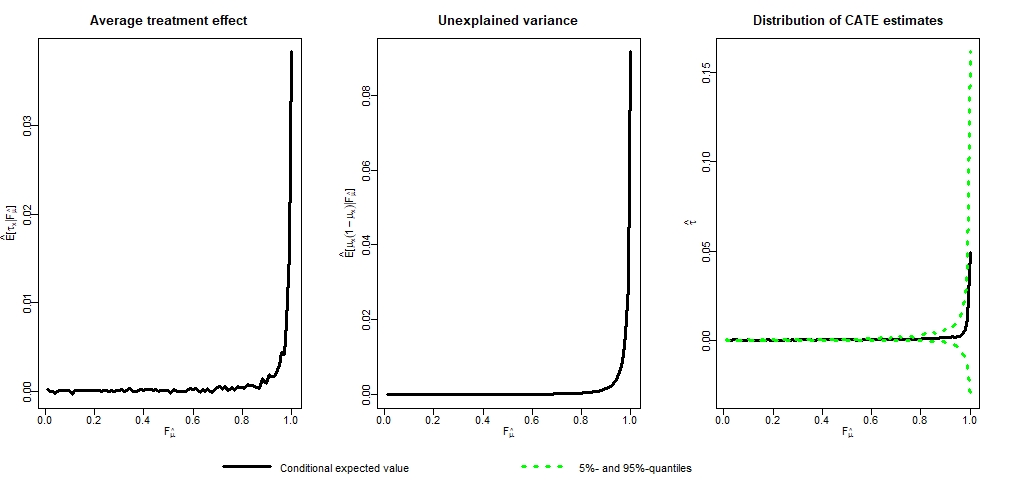}
 \captionof{figure}{Evidence of the heteroskedasticity effect. The left and middle plot show the estimated average treatment effect respectively the estimated unexplained variance at each position in the ranking. The right plot shows the average estimates of the T-learner as well as the 5\% and 95\% quantiles at each position in the ranking.}
 \label{fig:criteo_hetero}
\end{figure}

Having found that the T-learner suffers from ranking bias, the question then becomes whether this ranking bias is really due to the heteroskedasticity effect, which we examined in this paper. As the Criteo data set is real-world data, we can neither observe the conditional outcome probability $\mu_x$ nor the model variance $Var[\hat{\mu}|x]$. So, we can not prove the heteroskedasticity effect in the same way as we did in the simulation. However, we can still provide sound evidence for the heteroskedasticity effect. To this end, we divide the observations of the test data set into 100 buckets according to the predictions $\hat{\mu}$ of the outcome model. The first group $\{x:F_{\hat{\mu}}(\hat{\mu}(x))\in [0,0.01]\}$ contains the 1\% individuals with the lowest predictions. The next group $\{x:F_{\hat{\mu}}(\hat{\mu}(x))\in [0.01,0.02]\}$ contains the individuals with predictions between the 1\%- and the 2\%-quantile, and so on. Within all of these groups, we estimated the average unexplained outcome variance and the average treatment effect. For the average unexplained variance in each bucket, we applied the following estimator
\begin{align*}
    \hat{E}[\mu_x(1-\mu_x)]=\frac{1}{N}\sum_{i=1}^{N}\hat{\mu}(x_i)(1-\hat{\mu}(x_i)).
\end{align*} For the average treatment effect within each bucket, we applied the estimator
\begin{align*}
    \hat{E}[\tau_x]=\frac{1}{N_w}\sum_{i=1}^{N_w}y_{i}|_{w=1}-\frac{1}{N_{\bar{w}}}\sum_{i=1}^{N_{\bar{w}}}y_{i}|_{w=0},
\end{align*} where $N_w$ and $N_{\bar{w}}$ refer to the number of treated respectively untreated individuals within the bucket. We also calculated the average value and the 5\% and 95\%-quantiles of the T-learner estimates within each bucket. The results are provided in figure \ref{fig:criteo_hetero}.   

It seems that there is little or no treatment effect in individuals with an estimated outcome probability below the 90\% quantile. In contrast, for the individuals with the highest predicted outcome probability, there seems to be a relatively high treatment effect. The plot for the estimated unexplained variance looks very similar. Appearently, the individuals with high treatment effect also tend to have the highest unexplained variance in their outcomes. The right plot shows the average as well as the 5\% and the 95\% quantiles of T-learner predictions within each bucket. We can see that the T-learner predicts in average the highest treatment effect for those individuals with the highest predicted outcome probability. This is plausible, since the left plot shows that these individuals tend to have the highest treatment effect. The gap between the 5\% and 95\% quantiles indicates the variances of T-learner estimates within each bucket. There is almost no gap between the quantiles for predicted outcome probabilities below the 90\% quantile. Then the gap widens with increasing predicted outcome probability. These results align perfectly with our theory that heteroskedasticity caused the ranking bias: Individuals with high treatment effects tend to have a high unexplained outcome variance and therefore a high uplift model variance. In consequence, they tend to be ranked at the top or bottom by the uplift model.

\section{Discussion}
In this paper, we discussed a potential problem in the application of uplift models, namely the impact of heteroskedasticity. We have shown, that there is a tendency of uplift models to rank those individuals with a high unexplained variance in the outcome at the bottom or top. In case the CATE tends to increase with the unexplained variance in the outcome, the heteroskedasticity causes uplift models to rank a significant share of individuals with high CATE at the bottom. This ranking bias poses a potential problem of the uplift modeling approach. 

We expect ranking bias to appear in many real-world applications. We have demonstrated ranking bias on the Criteo real-world data set. In addition to this empirical evidence, there is another good reason to assume that ranking bias will appear in many real-world applications: In most applications, for treatment prioritization, the rate of positive outcomes is far below 50\%: In online marketing, the conversion rate is typically low. For advertisement, the click-through rate is usually also low and for applications where the goal is to prevent churn, the churn rate is also mostly low. If the rate of positive outcomes is so low, the conditional outcome probabilities will almost always stay below 50\%. Hence, a growth in conditional outcome probability goes in hand with a growths of the unexplained outcome variance. At the same time, experience in the uplift domain shows, that individuals with a high conditional outcome probability also tend to have the highest treatment effects.\citep{fernandez2022causal} So, the situation that high unexplained outcome variance goes in hand with high CATE values is likely to appear often in practice. Hence, we would expect the problem of ranking bias by heteroskedasticity to be of high practical relevance.

One solution to the ranking bias problem, which we briefly discussed, would be to allocate treatment also to the lowest ranked individuals, if such ranking bias is diagnosed. This works, because we know that the ranking bias tends to push individuals with high treatment effects to the bottom (rather than anywhere in the middle of the ranking). Another solution could be to use an outcome model instead of an uplift model to prioritize treatment. Such an outcome model will prioritize the individuals high, which would be ranked at the bottom by the uplift model, due to their high magnitude of unexplained outcome variance. It would also be possible to use a combination of an outcome model and an uplift model to decide about the treatment allocation. There is room for further research about methods to deal with the heteroskedasticity effect.

Finally, it needs to be said that the heteroskedasticity effect does not only occur for binary outcome variables. Some studies in the uplift literature consider the case of limited distributed targets, like monetary amounts spend by customers, which can be zero or take various positive values.\citep{gubela2020response,haupt2022targeting,baier2022profit} For such non-binary outcome variables the impact of heteroskedasticity might even be stronger than for a binary outcome, because the amount of the outcome also influences the unexplained outcome variance. This leaves even more room for research about the effect of heteroskedasticity on uplift modeling.

\appendix

\section{Qini curve}\label{ap_auuc}
The expected value of the Qini curve at point $(1-p)\in[0,1]$ is given by
\begin{align*}
    f(1-p)=E[\tau_{x}|F_{\hat{\tau}}(\hat{\tau})\geq p]\cdot (1-p).
\end{align*} Now, we derive a formula for $f'(1-p)$. It holds
\begin{align*}
    &E[\tau_{x}|F_{\hat{\tau}}(\hat{\tau})\geq p-h]\cdot (1-p+h)\\
    &=E[\tau_{x}|F_{\hat{\tau}}(\hat{\tau})\geq p-h]\cdot P[F_{\hat{\tau}}(\hat{\tau})\geq p-h]\\&=E[\tau_{x}|F_{\hat{\tau}}(\hat{\tau})\geq p]\cdot P[F_{\hat{\tau}}(\hat{\tau})\geq p]+E\left[\tau_{x}|F_{\hat{\tau}}(\hat{\tau})\in [p-h,p]\right]\cdot P\left[F_{\hat{\tau}}(\hat{\tau})\in [p-h,p]\right]\\
    &=E[\tau_{x}|F_{\hat{\tau}}(\hat{\tau})\geq p]\cdot (1-p)+E\left[\tau_{x}|F_{\hat{\tau}}(\hat{\tau})\in [p-h,p]\right]\cdot h.
\end{align*} Hence, we obtain
\begin{align}
    f'(1-p)&=lim_{h\rightarrow 0}\frac{f(1-p+h)-f(1-p)}{h}\nonumber\\
    &=lim_{h\rightarrow 0}\frac{E[\tau_{x}|F_{\hat{\tau}}(\hat{\tau})\geq p-h]\cdot (1-p+h)-E[\tau_{x}|F_{\hat{\tau}}(\hat{\tau})\geq p]\cdot (1-p)}{h}\nonumber\\
    &=lim_{h\rightarrow 0}E\left[\tau_{x}|F_{\hat{\tau}}(\hat{\tau})\in [p-h,p]\right]\nonumber\\
    &=E\left[\tau_{x}|F_{\hat{\tau}}(\hat{\tau})=p\right]\label{eq_deriv_qini}.
\end{align}

\bibliographystyle{unsrtnat}
\bibliography{template}

\end{document}